\documentclass[balance,upint,subscriptcorrection,varvw,mathalfa=cal=boondoxo,spanish,french,vietnamese,russian,greek,pdf-a,colorlinks]{asmeconf}
\usepackage{lipsum}
\usepackage{epsfig} 
\usepackage{graphicx}
\usepackage{amsmath}
\usepackage[export]{adjustbox}
\usepackage{subcaption}
\usepackage[long]{optidef}
\usepackage{multirow}
\usepackage{hyperref}
\begin{document}

\ConfName{Proceedings of the ASME 2024\linebreak International Design Engineering Technical Conferences and 
Computers and Information in Engineering Conference}
\ConfAcronym{IDETC/CIE2024}
\ConfDate{August 25--28, 2024}
\ConfCity{Washington, DC}
\PaperNo{IDETC2024-143607}

\newcommand{\eg}{{\em e.g.}}
\newcommand{\ie}{{\em i.e.}}
\newcommand{\etc}{{\em etc.}}
\newcommand{\etal}{{\em et~al.}}

\hypersetup{
    pdftitle={Inverse Design With Conditional Cascaded Diffusion Models}
}

\title{Inverse Design With Conditional Cascaded Diffusion Models}

\SetAuthors{%
	Milad Habibi\affil{1},
	Mark Fuge\affil{1}\CorrespondingAuthor{fuge@umd.edu}
	}

\SetAffiliation{1}{Center for Risk and Reliability, Department of Mechanical Engineering, University of Maryland, College Park, MD}

\maketitle 




\begin{abstract}
Adjoint-based design optimizations are usually computationally expensive and those costs scale with resolution. To address this, researchers have proposed machine learning approaches for inverse design that can predict higher-resolution solutions from lower cost/resolution ones. Due to the recent success of diffusion models over traditional generative models, we extend the use of diffusion models for multi-resolution tasks by proposing the conditional cascaded diffusion model (cCDM). Compared to GANs, cCDM is more stable to train, and each diffusion model within the cCDM can be trained independently, thus each model's parameters can be tuned separately to maximize the performance of the pipeline. Our study compares cCDM against a cGAN model with transfer learning.

Our results demonstrate that the cCDM excels in capturing finer details, preserving volume fraction constraints, and minimizing compliance errors in multi-resolution tasks when a sufficient amount of high-resolution training data (more than 102 designs) is available. Furthermore, we explore the impact of training data size on the performance of both models. While both models show decreased performance with reduced high-resolution training data, the cCDM loses its superiority to the cGAN model with transfer learning when training data is limited (less than 102), and we show the break-even point for this transition. Also, we highlight that while the diffusion model may achieve better pixel-wise performance in both low-resolution and high-resolution scenarios, this does not necessarily guarantee that the model produces optimal compliance error or constraint satisfaction.

\end{abstract}

\section{Introduction}

Optimizing designs, especially in the context of adjoint-based multi-physics shape and topology optimization (TO), can be a time-intensive process, often requiring costly iterations to achieve the desired designs~\cite{sigmund13}. For instance, one of the most commonly used methods to solve forward TO is Solid Isotropic Material with Penalization (SIMP) method~\cite{bendsoe1989,tcherniak2002}. This method is based on iterative optimization techniques over continuous, density-based methods to find the optimal material distribution. 
Despite the wide popularity of these methods, they suffer from two main pitfalls: (1) often requiring expensive simulations and (2) getting trapped in local optima due to their gradient-based solution approach. To overcome these challenges for TO problems, previous studies combine TO with inverse design methods driven by Machine learning. These methods complement adjoint methods by reducing the time and resources required to explore a broader range of design solutions compared to optimizing a single design with an iterative forward model. For instance, Habibi~\etal~\cite{habibi2023} showed that ML-based inverse design can predict designs with $95\%$ relative performance improvement for 2D heat conduction topology optimization problem compared to the SIMP method and more importantly it can accelerate the topology optimization even with some small amount of training data.

One of the main uses of ML-based design methods is addressing TO computational costs which scale with resolution. Several key questions arise in this context, including identifying techniques and models that can effectively reduce costs, determining the volume of data necessary for optimal model performance, and finding the right balance between gathering examples of low and high resolution and dimensions. A common technique to answer this challenge in the community is producing high-resolution images from low-resolution data which is also known as super-resolution~\cite{park2003}. Some researchers used Generative Adversarial Networks (GANs) trained on low-resolution TO datasets to perform upscaling to produce high-resolution images. For instance, Yu~\etal~implemented a two-stage refinement conditional GAN (cGAN) to predict optimal high-resolution structures (128x128), leveraging training on 64,000 low-resolution data samples (32x32) and corresponding high-resolution counterparts~\cite{yu2019}.  Additionally, other investigators have proposed leveraging transfer learning in conjunction with GANs or variational autoencoders (VAEs) for super-resolution. Behzadi~\etal, for example, used transfer learning and conditional GANs to fine-tune a model on 11,000 fixed volume fraction low-resolution samples (80x40) and 1,500 high-resolution samples (up to 200x400), contributing valuable insights to the delicate balance between low and high-resolution examples in the transfer learning context~\cite{behzadi2022}. 

In this study, we aim to extend Maz~\etal's comparison of the performance and training simplicity of condition diffusion models versus that of GANS in TO~\cite{maze2023diffusion}, but in the context of multi-resolution problems. Our primary objective is to quantify the extent to which diffusion-based generative models surpass their GAN-based counterparts in addressing low to high-resolution super-resolution challenges. Additionally, we explore the performance of each model under fixed-cost considerations, systematically examining their efficacy in handling both seen and unseen boundary conditions.
Specifically, the contributions of this paper are as follows:
\begin{enumerate}
\item We describe a conditional cascaded diffusion model-based framework for solving low to high-resolution engineering design problems.
\item We demonstrate the extent to which cascaded diffusion-based generative models outperform GAN-based generative models on low to high-resolution inverse design problems.
\item For a given threshold of performance (e.g., constraint violation, compliance, etc.), we study the comparative sample efficiency of diffusion models versus GAN models for low to high-resolution problems, noting a critical high-resolution sample threshold below which the cGAN model outperforms the diffusion model, and vice versa.
\end{enumerate}

\section{Related Works}
\subsection{Inverse Design and Topology Optimization}
Inverse design (ID) involves leveraging machine learning techniques to predict the optimal design, such as an airfoil mesh, based on specified input conditions, such as a desired Reynolds or Mach number~\cite{chen2021}. ID studies focus on the mapping between the input conditions and optimal design which can be used in place of an existing optimizer or as an initial guess for further optimization (warmstart optimization)~\cite{chen2021}. In the field of Mechanical engineering, ID has been explored in diverse applications,  designing and characterizing materials~\cite{lee2019,chen2021,kim2020,kim2020GEN,challapalli2021}, airfoil shape design~\cite{chen2021,sekar2019}, and topology optimization~\cite{habibi2023,bernard22,giannone23}.

Topology optimization (TO) finds the optimal distribution of the material in the design domain to optimize one or more objective functions while satisfying some constraints. Traditional TO methods rely heavily on Finite element analysis~\cite{sigmund13}. One of the most widely used formulations of TO problems are pseudo-density-based approaches such as the Solid Isotropic Material with Penalization method (SIMP)~\cite{bendsoe1989,bendsoe1988}. The goal of structural TO is to identify the optimal material distribution within the design domain to minimize structural deformation, subject to a volume constraint and prescribed load condition~\cite{sigmund2001}.

The specific minimization problem is formulated as follows:

\begin{mini}|l|
  {x}{c(x)=U^{T}KU}{}{}
  \addConstraint{\frac{V(x)}{V_{0}}}{=f}{}
  \addConstraint{KU}{=F}{}
  \addConstraint{0\le}{x_{min}}{\le1}
 \end{mini}
where $c(x)$ represents structural compliance, $U$ denotes global displacement, $F$ is the force vector, and $K$ is the global stiffness matrix. Additionally, $V(x)$ and $V_{0}$ represent material volume and design domain volume, respectively, with $f$ representing the prescribed volume fraction. The vector $x_{min}$ is introduced to prevent the singularity condition by specifying minimum relative densities.

Traditional TO methods usually suffer from two limitations: (1)~often requiring costly iterations and (2)~getting trapped in local optima due to their gradient-based
solution approach. To address these limitations of traditional TO many researchers developed various deep-learning methods~\cite{regenwetter2022,sosnovik2019,rawat2019}. For instance, Sharpe and Seepersad introduce conditional generative adversarial networks (cGANs) to generate a compact latent representation of structures resulting from traditional TO methods~\cite{sharpe2019}. Furthermore, Guo~\etal~ used variational autoencoder and style transfer to enhance topology optimization capabilities~\cite{guo2018}. Additionally, Nie~\etal~expand the generalizability of using the condition generative model by introducing TopologyGAN for more variant physical and boundary conditions~\cite{nie2021}.  Moreover, some researchers incorporate physics and generative models to solve the TO problems~\cite{challapalli2021,giannone2023,cang2019}. Additionally, reinforcement learning-based generative design frameworks have been explored by some researchers to improve the diversity of topology-optimized designs~\cite{jang2022,khan2022}.

The aforementioned works showcase diverse approaches to incorporating deep learning into design optimization. All are either supervised or unsupervised algorithms that rely on the use of enough amount of training data to predict acceptable results. But how much of this expensive high-resolution data is enough? There are a few papers that consider the limitation of high-resolution training data amounts~\cite{behzadi2021,behzadi2022}, typically leveraging transfer learning. In this paper, we aim to quantify and understand how training size affects ML-based inverse design model performance, but beyond the GAN-based models studied previously.

\subsection{Generative models for Inverse Design using Super-Resolution}
One of the main challenges in solving inverse design problems, especially TO problems is the computational cost which scales with the resolution. To tackle this issue, researchers often refer to super-resolution techniques from computer vision~\cite{farsiu2004,regenwetter2022}. Yu~\etal~\cite{yu2019} used a two-stage refinement, employing a conditional generative adversarial network (GAN) for upscaling and predicting near-optimal structures, while Li~\etal~employed a Super-Resolution GAN (SRGAN) for refining high-resolution structures~\cite{li2019}.

While various GANs and Variational Autoencoders (VAEs) have been applied in these methods, an alternative approach involves transfer learning, a technique transferring knowledge from one domain to another effectively~\cite{zhuang2020,weiss2016}. Behzadi and Ilieș introduced a transfer learning method based on a convolutional neural network (CNN) to the topology optimization community~\cite{behzadi2021}. Their approach involved training a feed-forward CNN on low-resolution data, transferring knowledge by locking model weights, and adding layers for upscaling output resolution. Subsequent work by Behzadi and Ilieș proposed a design exploration framework leveraging transfer learning and conditional GANs, surpassing previous generalization capabilities~\cite{behzadi2022}.

Dhariwal and Nichol demonstrated that diffusion models excel in image generation, outperforming GANs, and offer easier training and adaptability to various tasks~\cite{dhariwal2021}. Building on this, Maz{\'e} and Ahmed showcased that a topology-based diffusion model significantly outperforms state-of-the-art conditional GANs, reducing average errors in physical performance by a factor of eight and producing fewer infeasible samples~\cite{maze2023diffusion}. Inspired by these findings, our study addresses the different problem of super-resolution in diffusion models, and answers the question ``to what extent do diffusion-based generative models surpass GAN-based counterparts in low to high-resolution problems, and is the adoption of diffusion models appropriate for super-resolution tasks?''
\subsection{Diffusion Models}
Diffusion models, as introduced by Sohl-Dickstein~\etal~\cite{sohl2015}, represent a class of deep generative models. These models include two main components: the forward and backward diffusion processes.
The forward diffusion process entails the addition of noise to sample vectors over $T$ steps to a sample from data distribution $x_{0} \backsim q(x_{0})$, generating a sequence of noisy samples $x_{1}, x_{2}, ..., x_{T}$, following the Markov chain:
\begin{equation} \label{eq1}
q(x_t | x_{t-1}) = \mathcal{N}(x_t ; \sqrt{1 - \beta_{t}} x_{t-1}, \beta_{t} I)
\end{equation}
Where $q(x_t | x_{t-1})$ the forward process posteriors, $\beta_{t}$ is variance schedule at time $t$. As the forward diffusion progresses, the initial data sample $x_{0}$ gradually loses its features, a phenomenon controlled by the variance schedule denoted as ${(\beta_{t})_{t=1}^{T} \in (0, 1)}$.

The reverse process seeks to approximate the true posterior using a parametric Gaussian process. This involves recreating the true sample from a Gaussian noise input, expressed as
\begin{equation}
p_\theta(x_{t-1} | x_t) = \mathcal{N}(\mu_\theta(x_t), \Sigma_\theta(x_t))    
\end{equation}
In the reverse process the new image is generated by sampling a data from pure noise and removing noise gradually using reverse process equation. The reverse process involves training the neural network model to predict the mean and variance of the denoising process. In our model we assumed variance of the denoising process $\Sigma_\theta$ is fixed and constant and we learn the mean variable $\mu_\theta$~\cite{ho2020}. In summary, the whole intuition behind the diffusion model is training models to reverse a noise process and mapping a Gaussian noise distribution to the data distribution. Readers interested in finding out more mathematical details about diffusion models are directed to \cite{luo2022}.

Diffusion models, as demonstrated by Ho~\etal~\cite{ho2020}, show superior image quality and greater stability compared to GANs. While Denosing Diffusion Models (DDMs) are relatively novel in the context of engineering design, there have been notable studies exploring their application in creating two-dimensional structures. One noteworthy example is the work of Maz{\'e} and Ahmad~\cite{maze2023diffusion}, who introduced a condition diffusion model-based architecture, TopoDiff. This innovative approach was employed for performance and manufacturability-aware topology optimization, outperforming state-of-the-art conditional GANs by reducing the average physical performance by a factor of eight. Building on this advancement, Giannone~\etal~\cite{giannone2023} proposed a learning diffusion model framework that eliminates the need for expensive finite element preprocessing, resulting in improved design performance. 
These studies highlight diverse methodologies for integrating diffusion models into design optimization, yet a crucial aspect remains unexplored. None of these works have delved into the potential of super-resolution within diffusion models as compared to GANs for engineering designs which we aim to explore in this study.

\section{Methodology}
To address the contributions outlined in the introduction we divide the methodology into the following sections: (1) introducing the cascaded pipeline, (2) what are the diffusion model parameters and its architecture, (3) what cGAN model we bench-marked against our proposed diffusion model, and (4) how we measure and evaluate our models.
\subsection{Conditioned Cascading Pipelines}
\begin{figure*}
\centering
\includegraphics[width=\linewidth]{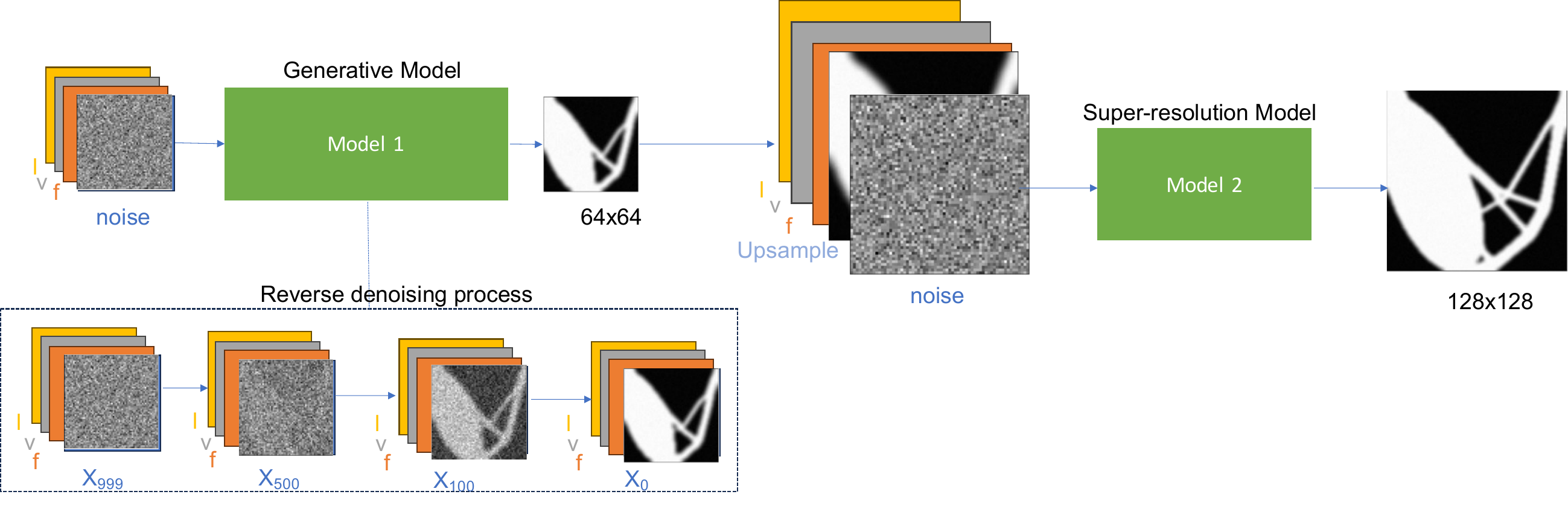}
\caption{Proposed cCDM pipeline for TO super-resolution. Model 1 is a standard conditional diffusion model for TO. L is the load applied, represented with an arrow on the topology; v
is the volume fraction; f represents the physical field including the stress and strain field. Model 2 is a super-resolution model conditioned on boundary conditions and upsampled prediction of Model 1. }
\label{cascaded.ps}
\end{figure*}
Inspired by previous studies~\cite{ho2022,saharia2022}, we introduce a framework for addressing multi-resolution engineering challenges, illustrated in Fig~\ref{cascaded.ps}. Our core approach lies in conditional cascaded diffusion (cCDM), which involves a sequence of generative diffusion models. At the lower resolution, we use a generative diffusion model conditioned on specified boundary conditions. Following the methodology of prior works~\cite{maze2023diffusion, nie2021}, we perform channel-wise concatenation with channels representing the volume fraction, strain energy density, von Mises stress, and applied loads on the domain's boundary. These conditionally augmented features are then integrated with the low-resolution image. Moving to higher resolutions, our framework incorporates a super-resolution diffusion model conditioned by channel-wise concatenation. This concatenation involves the low-resolution image predicted by the preceding diffusion model, which is further processed through bilinear upsampling to achieve the desired resolution. Also, the conditioning at the high resolution includes the given boundary conditions. One important advantage of employing these cascading models is their independence during training, allowing for separate tuning of hyperparameters at each stage. This flexibility improves the overall performance of the pipeline, as parameters can be optimized individually for each resolution level. We describe the specific network architectures and diffusion parameters in later sections.

\subsection{ U-Net Architecture and Diffusion Parameters}
We used a UNET architecture as the basis for our diffusion models implementation, obtained from the following publicly available repository~\cite{dome272}. We made specific modifications based on our needs, notably, we removed the attention block within the UNET architecture to reduce the computational complexity and because including the attention block worsened performance in our initial tests. 
For the baseline model, we used a linear variance $\beta$ schedule lineary increasing from $10^{-4}$ to $0.02$ as it is suggested in the previous study~\cite{ho2020} with a total of 1000 steps where the 1st step represents the lowest amount of noise, and 1000 is the highest. 
\subsection{Conditional Generative Adversarial Networks and Transfer Learning }
We adapted the Conditional Generative Adversarial Network (cGAN) architecture from Behzadi~\etal~\cite{behzadi2022}, sourced from their public GitHub repository, to benchmark against our cCDM Our modifications prioritized improving stability and performance over our dataset. To address challenges encountered during model training we used a Wasserstein loss, which reduced mode collapse issues we found during testing relative to the original implementation in~\cite{behzadi2022}.

\subsection{Dataset}
To test our models and be comparable to the existing state-of-the-art we created a dataset via a classical structure compliance topology optimization problem based on the SIMP approach using the code provided by Andreassen~\etal~which we detail below~\cite{andreassen2011}. This problem aims to minimize the structural compliance of a beam (represented by a square domain) while satisfying the constraints on material volume and boundary conditions, including force magnitude, location, and direction.

\begin{figure}
\centering
\includegraphics[width=\linewidth]{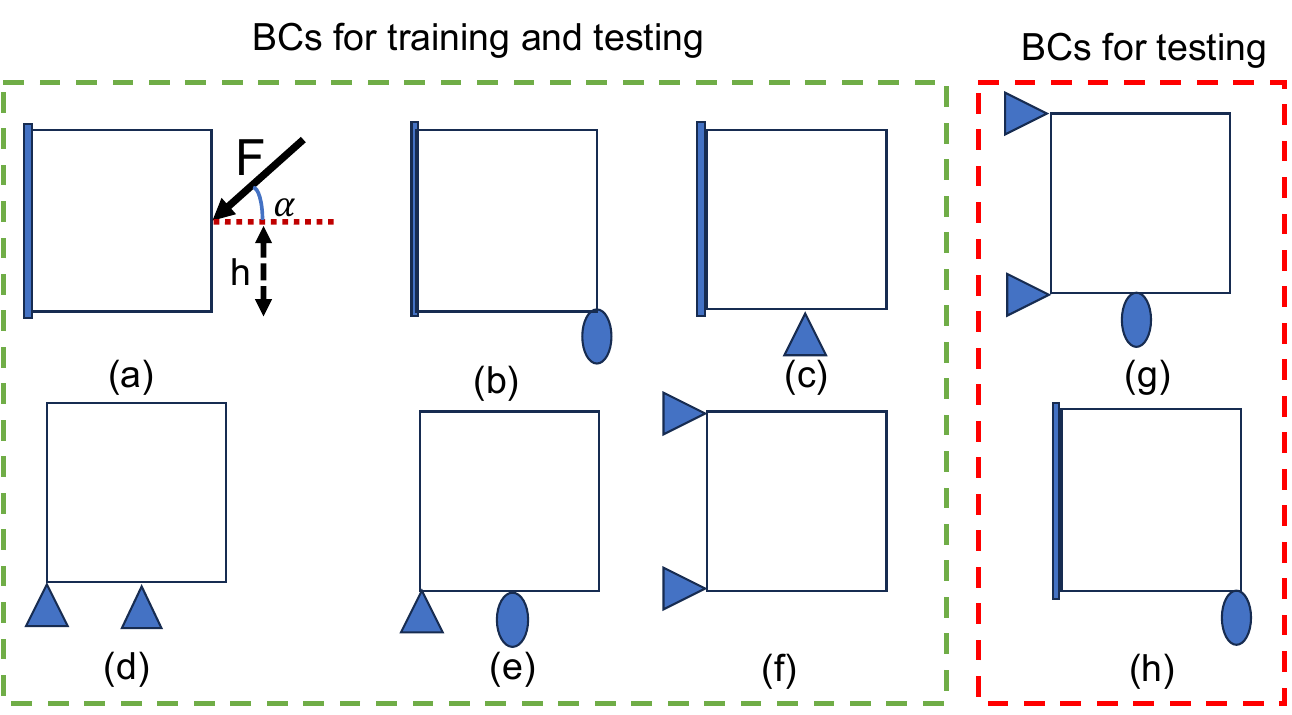}
\caption{The physical layout of topology optimization problem of beams and corresponding training and testing boundary condition. Design domain of a cantilever beam with design parameters of force location (h) and direction ($\alpha$) on the right side of the beam. Boundary conditions: (a)-(f) are used for training and testing and (g)–(h) are unseen BCs only used for testing.}
\label{fig_ph.ps}
\end{figure}

Consistent with past work, we selected two sets of boundary conditions, as illustrated in Fig. \ref{fig_ph.ps}(a), using the first set for both training and testing, while the second set served as unseen boundary conditions exclusively for testing purposes. The force magnitude was set to 5000N, and the design domains included volume fractions ($v$) ranging from 0.30 to 0.60, force locations ($h$) ranging from 0 to 1, and force directions ($\alpha$) on the right side of the design domain. 

For the seen boundary conditions, we divided the volume fraction range into 20 segments and the force location and direction ranges into 10 segments, resulting in 2000 designs for each boundary condition. For the unseen boundary conditions, we used 10 segmentations for volume fraction and force location and direction, resulting in 1000 designs for each boundary condition. Within the seen boundary conditions category, we employed a random selection process to allocate 200 from each seen boundary condition designs for testing purposes, while the 1500 designs were dedicated to training the model which resulted in 9000 training data and 1200 testing seen boundary conditions. Similarly, for the unseen boundary conditions category, we randomly selected 1000 designs for testing, with an equal distribution of designs across different unseen boundary conditions. These unseen boundary conditions served as a critical benchmark for evaluating the model's ability to test the generalizability of the models. Also, we generated the dataset at two distinct resolutions: a low resolution of 64x64 and a high resolution of 128x128. 

Inspired by previous studies~\cite{nie2021,maze2023diffusion} highlighting the improved prediction accuracy achieved by incorporating physical fields such as von Mises stress and strain energy density instead of explicitly representing boundary conditions, we introduced these fields as additional input channels in the generative models. Additionally, force and volume fraction channels were included to enable the network to identify relationships between physical boundary conditions and fields, focusing on structural patterns that serve as effective load paths.

The physical representation dataset comprises five channels:
\begin{itemize}
\item The first channel is uniform and contains the volume fraction constraint;
\item The second channel includes the von Mises stress values  of the design domain under the given constraints;
\item The third channel is the strain energy density of the design considering both load constraints and boundary conditions. Strain energy is defined as $W=\frac{\sigma_{x}\epsilon_{x}+\sigma_{y}\epsilon_{y}+2\sigma_{xy}\epsilon_{xy}}{2}$; 
\item The fourth channel represents the load constraints in the x-direction. Each node indicates the force applied in the x-direction direction (or 0 if there's no force applied);
\item The fifth channel represents the load constraints in the y-direction. Each node indicates the force applied in the y-direction direction (or 0 if there's no force applied);
\end{itemize}
Since our study involved the comparison of different models, variations existed in the input data for each model. Specifically, the GAN model was trained on the physical representation dataset, while the diffusion model on low resolution incorporated an extra channel in the black-and-white image, representing the optimal topology. Furthermore, the diffusion model on higher resolution included two extra channels, incorporating topology optimization on higher resolutions and upsampled topology predictions from the low-resolution diffusion model.

\subsection{Evaluation Metrics}
In this study, we use pixel-wise error as a final evaluation metrics as well as physical objectives since in designing generative models minimizing objective values is more important than classical computer vision metrics.
As result, we define and use the following three metrics for our case studies evaluations:
\begin{itemize}
    \item Mean Squared Error (MSE), which is defined as:
\begin{equation}
    MSE = \frac{1}{nm} \sum_{i=1}^{n} \sum_{j=1}^{m} \left( {y}_{ij}^{pred} - y_{ij}^{true} \right)^2,
\end{equation}
where \( n \) and \( m \) are the number of rows and columns, ${y}_{ij}^{pred}$ and $y_{ij}^{true}$ are the predicted and reference values of the element located in the \( i \)th row and \( j \)th column, respectively.

\item Volume fraction error (VFE) relative to the input volume fraction is defined as:
\begin{equation}
    VFE = \frac{|VF({y^{pred}}) - VF(y^{true})|}{VF(y^{true})}
\end{equation}
where $VF(y^{true})$ and  $VF(y^{pred})$ represent, respectively, the prescribed volume fraction and the volume fraction of the topology generated by our diffusion model.
\item Compliance error (CE) in relation to the ground truth is defined as:
\begin{equation}
    CE = \frac{C({y^{pred}_{f}}) - C(y^{true}_{f})}{C(y^{true}_{f})}
\end{equation}
where \( C(y^{true}_{f}) \) and \( C(y^{pred}_{f}) \) represent the compliance of the feasible topology generated by the SIMP method and the feasible topology generated by the generative models which do not violate the volume fraction constraints, respectively. It is important to note that a negative compliance error indicates that generative models produce a topology with lower compliance than the ground truth. Furthermore, we considered topologies that satisfy volume fraction constraints and can be considered feasible designs.

\end{itemize}

\section{EXPERIMENTAL RESULTS AND DISCUSSION}

\subsection{How well do different models perform in low-resolution?}
To assess the performance of the diffusion model and the cGAN model for low-resolution topology optimization, we used the two test sets described in Section 3.4. Our evaluation focuses on comparing the models' performance using pixel-wise error metrics as well as physical values.

\begin{figure}
\centering
\includegraphics[width=\linewidth]{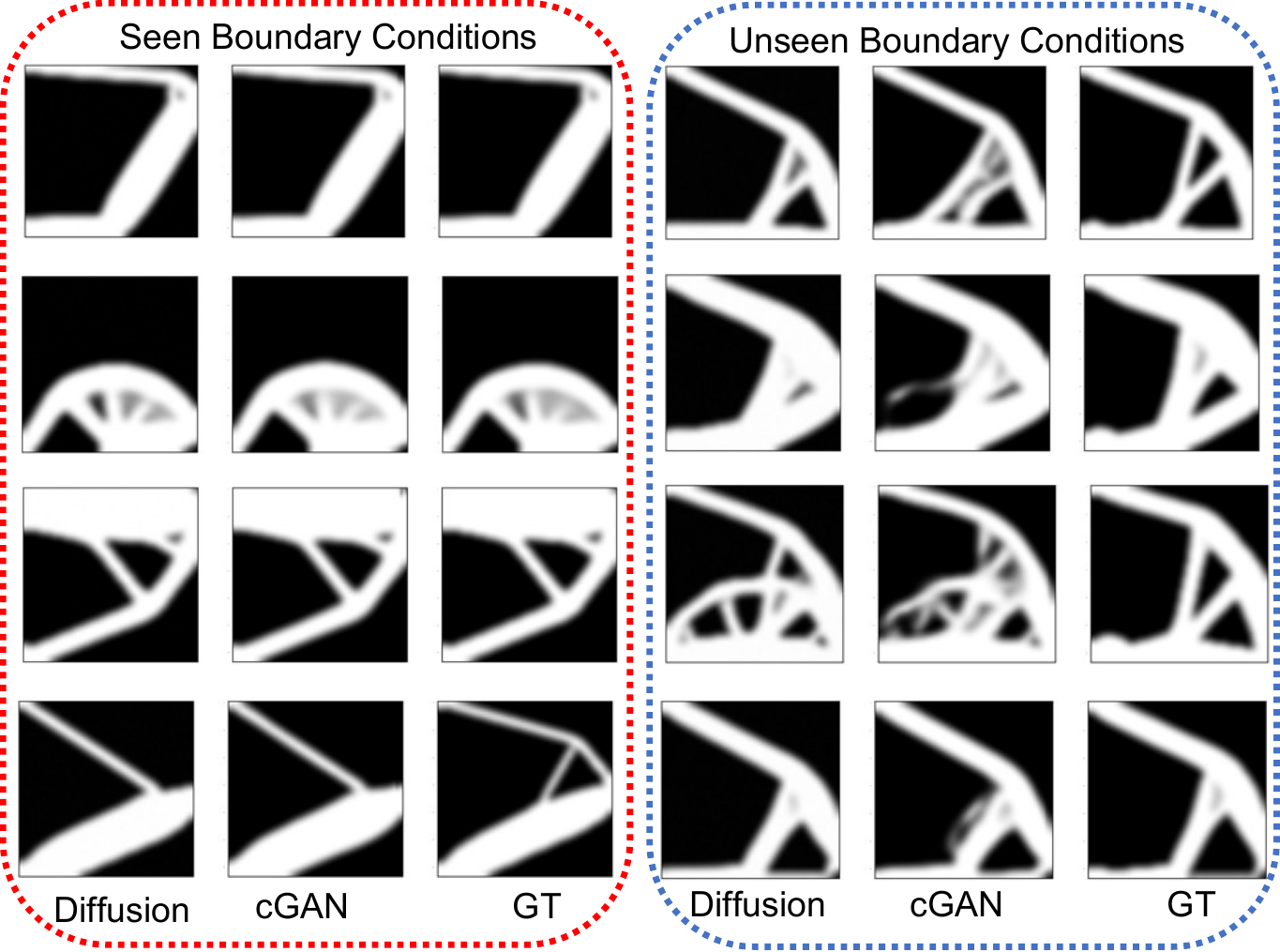}
\caption{ Comparison of generated structures on randomly selected samples from seens and unseen boundary conditions in the low-resolution data. The GT column represents the ground truth generated by the SIMP method.}
\label{LOW_RES.ps}
\end{figure}

Figure~\ref{LOW_RES.ps} shows examples of a few structures obtained with (1) the conditional GAN, (2) the conditional diffusion model, and (3) the SIMP method, which we label as the ``Ground Truth'' (GT), for randomly selected constraints from seen and unseen boundary conditions. As expected, the diffusion model exhibits superior performance qualitatively in terms of pixel-wise metrics in comparison to the cGAN model, aligning with previous research that showed diffusion models can outperform GANs on image synthesis~\cite{dhariwal2021}. However, our previous study has shown that optimizing solely for MSE values does not necessarily guarantee optimal design outcomes~\cite{bernard22}.

To quantitatively validate this observation, we summarize the performance of the generated structures across seen and unseen boundary conditions in Table~\ref{table:1}. Our analysis of the MSE indicates that the diffusion model outperforms the cGAN model in capturing finer details and minimizing pixel-wise errors across both seen and unseen boundary conditions. Interestingly, both models exhibit comparable performance in maintaining volume fraction constraints.
In addition, the diffusion model outperforms the cGAN model in minimizing compliance for more difficult test cases (unseen boundary condition test cases) by producing smaller median compliance errors. It should be noted we used the median for CE values since the median is less sensitive to outliers. This indicates that the topologies generated by the diffusion model are producing less structural compliance, highlighting its efficacy in producing feasible optimal designs.

\begin{table*}[htbp]
    \centering
    
    \begin{tabular}{lcccc}
        \toprule
        \text{Model} & \text{Boundary Condition} & \text{Average MSE ($\times 10^{-4}$)} & \text{Average VFE ($\times 10^{-4}$)} & \text{Median CE ($\times 10^{-2}$)} \\
        \midrule

        \multirow{2}{*}{Diffusion Model} & Seen & $\mathbf{6.67  \pm 2.00 }$ & $\mathbf{81.67  \pm 4.02 }$ & $0.34 \pm 0.04 $ \\
         & Unseen & $\mathbf{324.94 \pm 19.12}$ & $448.83  \pm 30.60$ & $\mathbf{4.18 \pm 8.54 }$ \\
        \midrule
        \multirow{2}{*}{cGAN Model} & Seen & $13.58  \pm 2.28 $ & $183.31 \pm 11.13$ & $\mathbf{0.09  \pm 0.15}$ \\
         & Unseen & $327.15 \pm 17.24 $ & $\mathbf{421.37  \pm 24.11 }$ & $5.86  \pm 425.85$ \\
        \bottomrule
        
    \end{tabular}
    
\caption{Performance Metrics Comparison in Low Resolution (64X64) on seen and unseen boundary conditions. Values after $\pm$ indicate the 95 \% confidence interval around averages/medians. The values in bold feature are the best ones for each boundary condition.}
\label{table:1}
\end{table*}

\subsection{How well do different models perform in high-resolution with limited training data?}

\begin{figure}
\centering
\includegraphics[width=\linewidth]{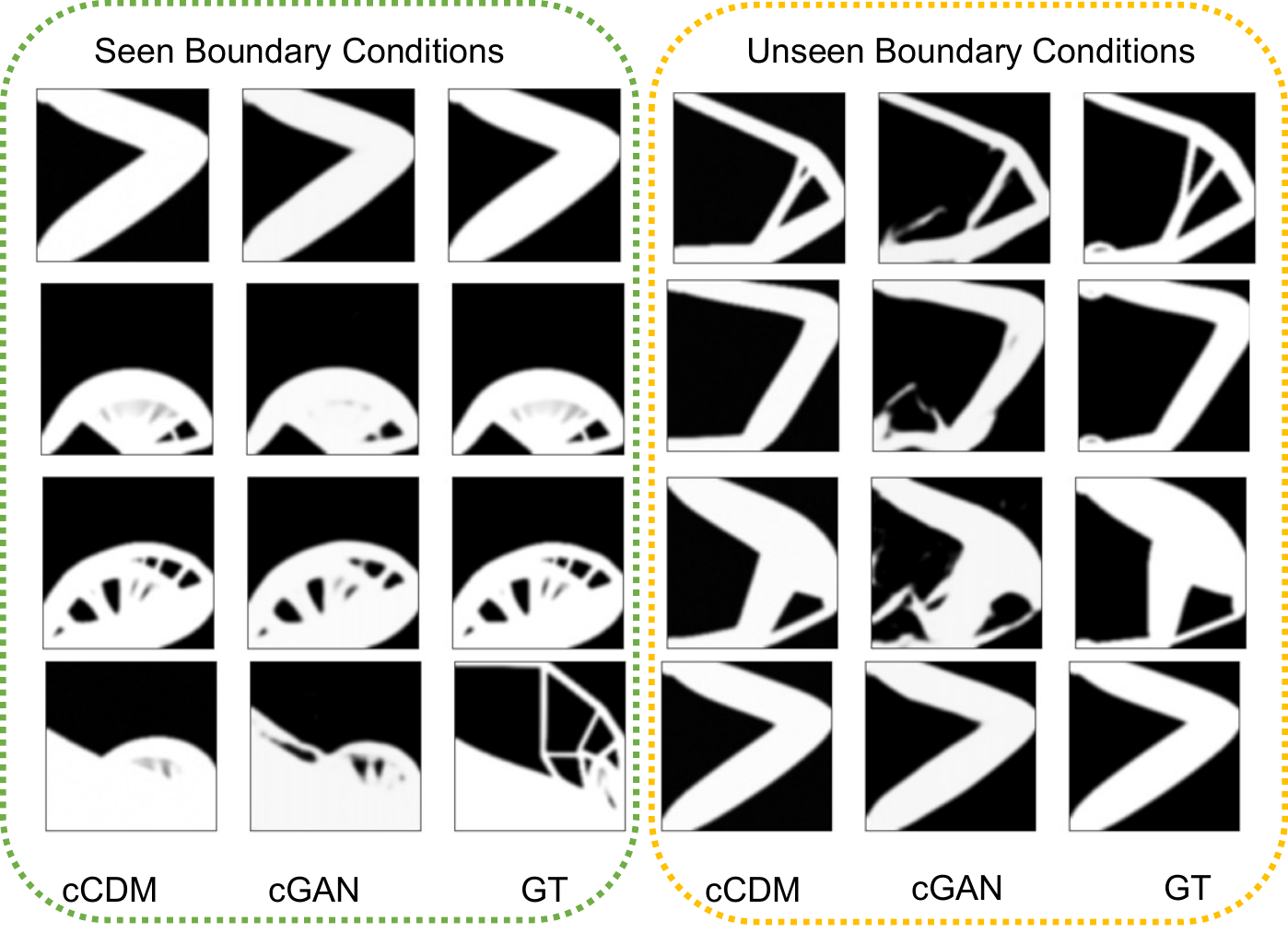}
\caption{ Comparison of generated structures on randomly
selected samples from seens and unseen boundary conditions in the high-resolution data. The GT column represent the ground truth generated by SIMP method.}
\label{HIGH_RES.ps}
\end{figure}

We evaluated the performance of the cCDM and the transfer of knowledge from low-resolution to high-resolution in the cGAN model. For this evaluation, we trained the cCDM using a limited number of training data (350 samples) from each seen boundary condition, resulting in a total of 2100 training samples for high-resolution. To provide a benchmark, we fine-tuned the cGAN model with transfer learning using the same training dataset.

Figure~\ref{HIGH_RES.ps} showcases examples of structures generated by the diffusion model and the cGAN model, along with SIMP method for randomly selected constraints from both seen and unseen boundary conditions. Consistent with expectations, qualitative analysis reveals the superior performance of the cCDM in terms of pixel-wise metrics compared to the cGAN model with transfer learning across both seen and unseen boundary condition test cases.

To quantify these comparisons, Table~\ref{table:2} provides a comprehensive overview of our evaluation metrics. The results clearly demonstrate the diffusion model's superiority over the cGAN model in all evaluated metrics for high-resolution topology optimization on both seen and unseen boundary conditions. Specifically, the diffusion model achieves lower MSE, VFE, and CE values, indicating its superior ability to capture finer details, preserve volume fraction constraints, and minimize compliance errors in super-resolution task.

\begin{table*}[htbp]
    \centering
    
    \begin{tabular}{lcccc}
        \toprule
        \text{Model} & \text{Boundary Condition} & \text{Average MSE ($\times 10^{-4}$)} & \text{Average VFE ($\times 10^{-4}$)} & \text{Median CE ($\times 10^{-2}$)} \\
        \midrule

        \multirow{2}{*}{cCDM} & Seen & $\mathbf{44.96  \pm 6.36 }$ & $\mathbf{99.11  \pm 5.32 }$ & $\mathbf{0.39  \pm 0.20 }$ \\
         & Unseen & $\mathbf{465.14  \pm 23.57}$ & $\mathbf{464.65  \pm 29.27 }$ & $\mathbf{10.44  \pm 5.40} $ \\
        \midrule
        \multirow{2}{*}{cGAN Model} & Seen & $98.09  \pm 8.52$ & $210.54 \pm 11.30$ & $3.20 \pm 60533.52$ \\
         & Unseen & $579.00  \pm 21.78 $ & $583.75  \pm 29.70 \times 10^{-3}$ & $17.29 \pm 20.30$ \\
        \bottomrule
        
    \end{tabular}
\caption{Performance Metrics Comparison in High Resolution (128X128) on seen and unseen boundary conditions. Values after $\pm$ indicate the 95 \% confidence interval around averages/median}
\label{table:2}
\end{table*}

\subsection{How well do models perform in high-resolution as a function of training data size?}

\begin{figure}
\centering
\includegraphics[width=\linewidth]{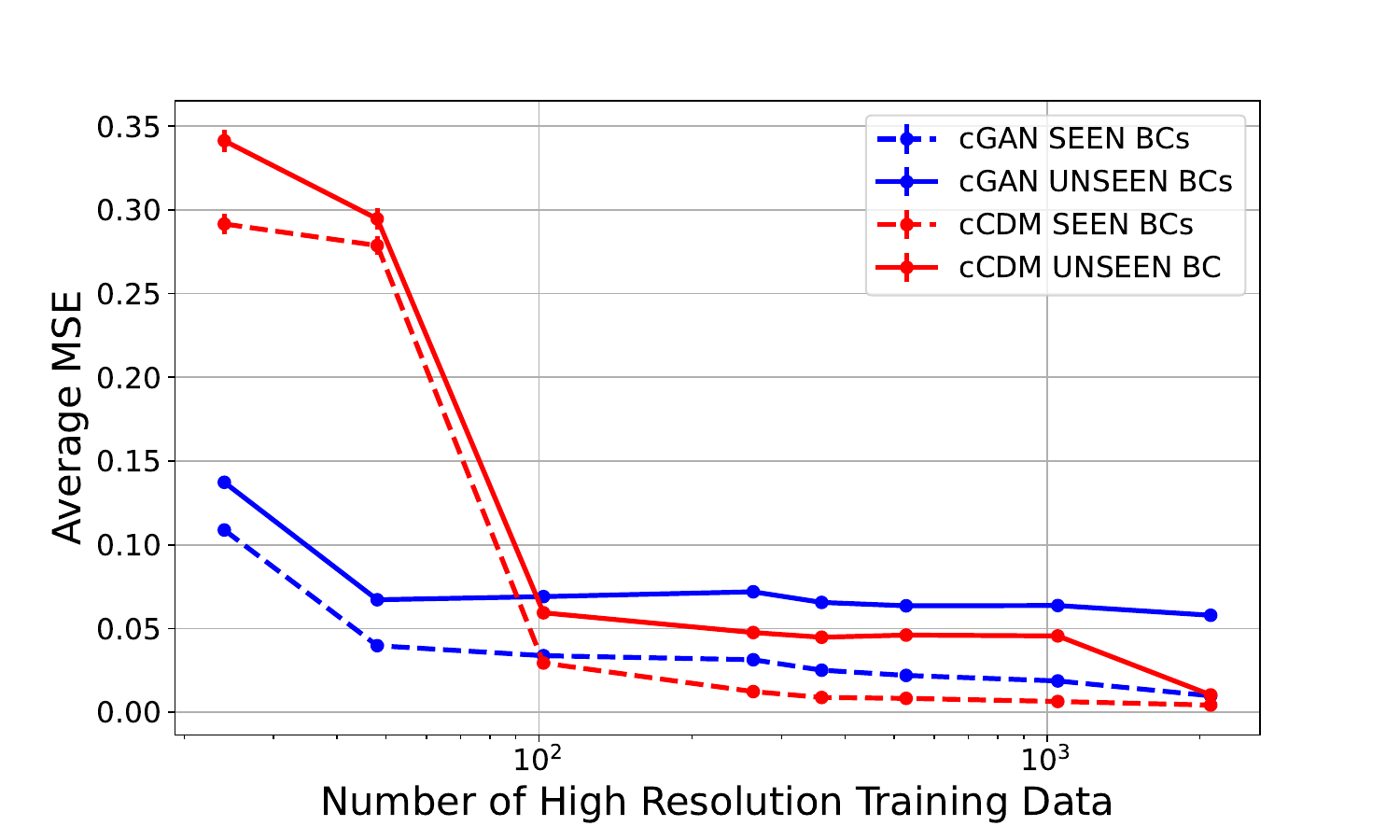}
\caption{ Mean MSE of the cGAN model with transfer learning versus the cascaded diffusion model across test data with seen and unseen boundary conditions as the number of high-resolution training data changes. Error bars represent the 95\% confidence intervals around the average MSE values, providing insights into the uncertainty of the measurements.}
\label{MSE.ps}
\end{figure}

\begin{figure}
\centering
\includegraphics[width=\linewidth]{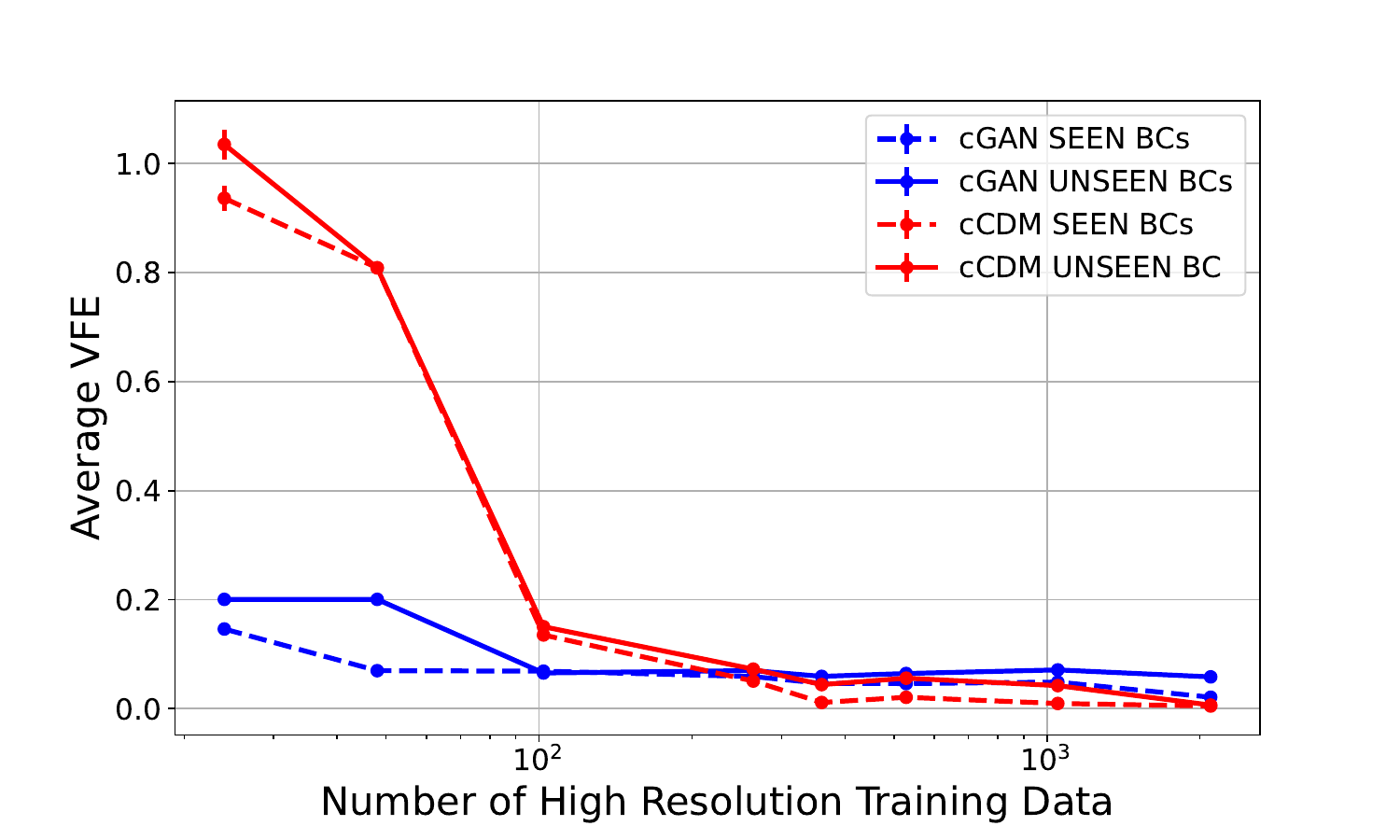}
\caption{ Mean of VFE of the cGAN model with transfer learning and cascaded diffusion models across test data with seen and unseen boundary conditions as the number of high-resolution training data decreases are showed. Error bars represent the 95\% confidence intervals around the average VFE values, providing insights into the uncertainty of the measurements.}
\label{VFE.ps}
\end{figure}

\begin{figure}
\centering
\includegraphics[width=\linewidth]{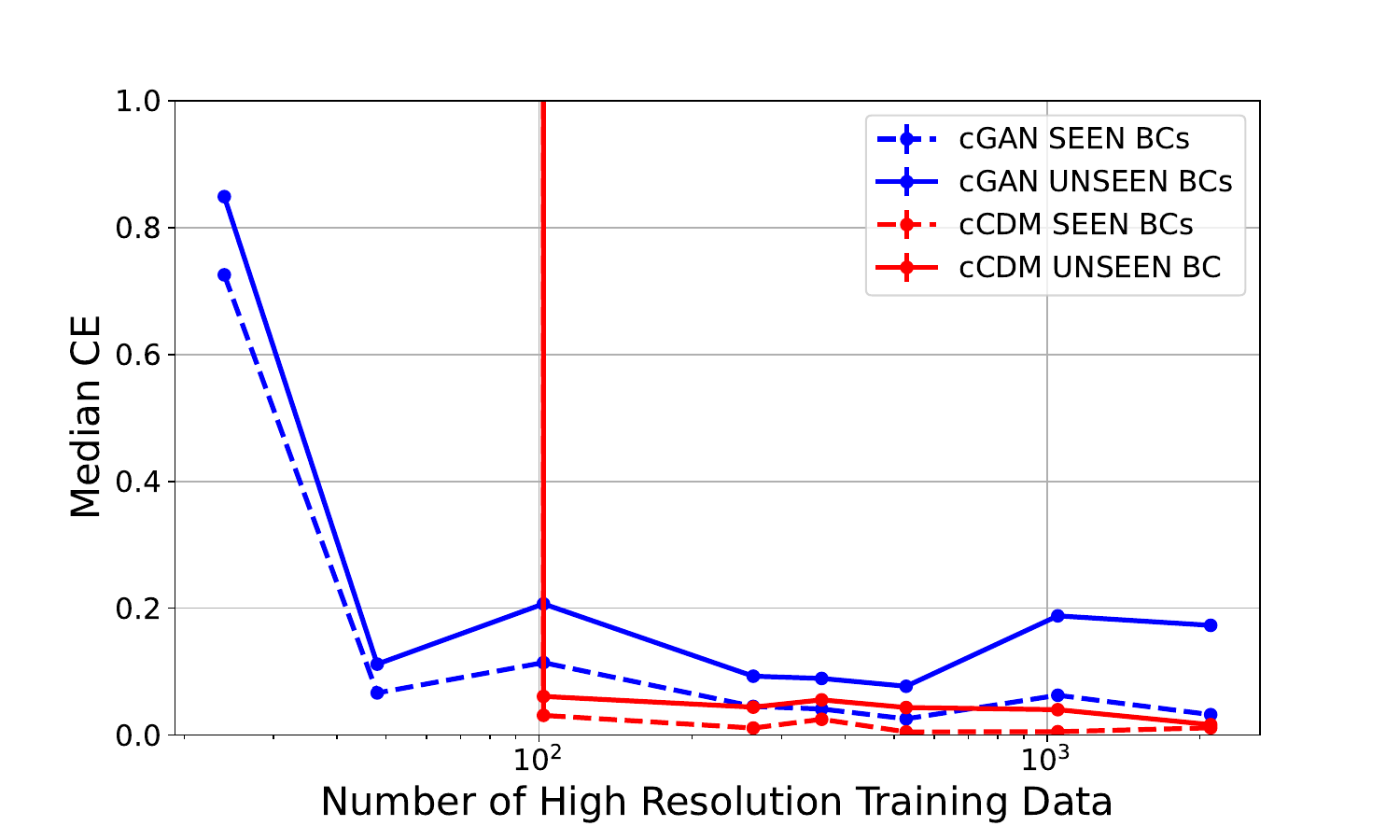}
\caption{ Median of CE of the cGAN model with transfer learning and cascaded diffusion models across test data with seen and unseen boundary conditions for feasible designs as the number of high-resolution training data decreases are showed.}
\label{CE.ps}
\end{figure}

One of the motivations to address this question stems from the high cost associated with generating high-resolution training data using the SIMP method. In response, we investigated the performance of the cGAN with transfer learning and cascaded diffusion models for high-resolution structures as we decreased the amount of high-resolution training data. The experiment aimed to assess how the performance of both models is impacted by the reduction in training data and to compare their adaptability to limited training resources, which can be the dominant cost. 

The number of training data varied from 24 to 2100. We excluded seen test boundary conditions and gradually reduced the size of the high-resolution training data. Each subsequent training set was obtained by randomly removing data from the next largest set. For example, the high-resolution dataset initially included 2100 designs, with 350 designs from each seen boundary condition. Subsequently, the dataset containing 1050 designs was created by removing 175 designs from each previously seen boundary conditions.

Figure~\ref{MSE.ps} illustrates the MSE performance of the cGAN with transfer learning and cCDM across seen and unseen boundary conditions as the number of high-resolution training data decreases. The cCDM outperformed the cGAN model, maintaining its superiority even as the amount of training data decreased for both sets of test data where the training dataset contains more than 102 samples. When the number of high-resolution data is limited (less than 102), however, we observe a reversal in performance, with the cGAN model with transfer learning outperforming the cascaded diffusion model.

Similarly, Figure~\ref{VFE.ps} shows how volume fraction error changes across both models as the number of training data decreases. Remarkably, the performance remains relatively stable up to 360 training data. However, beyond this threshold, a notable decrease in performance is observed, showcasing the sensitivity of both models to a low amount of training data. Notably, the cCDM consistently demonstrates superiority over the cGAN model, when the training dataset contains more than 102 samples. However, we observe a reversal in performance when the amount of high-resolution data is limited (less than 102), with the cGAN model with transfer learning outperforming the cascaded diffusion model under these conditions. 

Figure~\ref{CE.ps} illustrates the median values of CE since median value is less sensitive to outliers. We observed that the median value of the cCDM demonstrates a smoother trend, with a gradual increase as the number of training data decreases. It is worth noting that we only considered feasible designs that do not violate volume fraction constraints for compliance error calculation in Fig~\ref{CE.ps}. For completeness, we also compare the median CE across all designs (both including versus excluding infeasible designs) and provide this in the supplement material. Figure~\ref{CE.ps} highlights the consistent superiority of the diffusion model over the cGAN model when the training dataset contains more than 102 samples. However, similar to other metrics the reversal in performance is observed when the number of high-resolution data is limited (less than 102), with the cGAN model with transfer learning outperforming the cCDM under these conditions. Another important observation is that with only 48 designs, the median value of CE for the cascaded diffusion model is eight orders of magnitude higher than that of the cGAN model. Moreover, with 24 training data, the cascaded diffusion model failed to predict any feasible designs. However, as demonstrated in the supplementary material, considering all predicted designs (ignoring constraint violations) can lead to misleading conclusions, such as the best performance with a training size of 24.

\subsection{What is the effect of just the high-resolution model if we use a low-resolution oracle prediction?}

\begin{table*}[htbp]
    \centering
    
    \begin{tabular}{lcccc}
        \toprule
        \text{Model} & \text{Boundary Condition} & \text{Average MSE ($\times 10^{-4}$)} & \text{Average VFE ($\times 10^{-4}$)} & \text{Median CE ($\times 10^{-2}$)} \\
        \midrule

        \multirow{2}{*}{SR Model using SIMP } & Seen & $\mathbf{42.69  \pm 6.51 }$ & $\mathbf{49.16  \pm 2.87 }$ & $1.13  \pm 0.12$ \\
         & Unseen & $\mathbf{103.38  \pm 9.65}$ & $\mathbf{66.9  \pm 4.85 }$ & $\mathbf{1.62 \pm 0.26} $ \\
        \midrule
        \multirow{2}{*}{cCDM}& Seen & ${44.96  \pm 6.36 }$ & ${99.11  \pm 5.32 }$ & $\mathbf{0.39  \pm 0.20 }$ \\
         & Unseen & ${465.14  \pm 23.57}$ & ${464.65  \pm 29.27 }$ & ${10.44  \pm 5.40} $  \\
        \bottomrule
        
    \end{tabular}
\caption{Performance Metrics Comparison in High Resolution (128X128) on seen and unseen boundary conditions for model isolated generative model and cascaded model. Values after $\pm$ indicate the 95 \% confidence interval around averages/medians}
\label{table:3}
\end{table*}

While the above results show how both generative models perform (Low resolution plus subsequent High Resolution), it is possible that deficiencies in the low-resolution model could be impacting the overall results. In this section, we try to isolate the effect of just the super-resolution in cCDM by providing ground truth low-resolution samples generated by the adjoint solver (SIMP method) in place of the low-resolution design produced by the first stage of the conditional generative model\textemdash we refer to this setup as the ``Super Resolution (SR) Model using SIMP,'' in constrast to our joint cCDM model. The evaluation focuses on assessing whether the low-resolution generative model's accuracy influences the performance of the multi-resolution model. The motivation for this experiment is both scientific and practical: we wish to understand the isolated causal effect of the Super Resolution element of the pipeline, and in practice, it might be cheap to generate good SIMP TO solutions such that the ML methods may only be needed for higher resolution predictions. 

Table~\ref{table:3} compares the error metrics under both scenarios and shows that when using the upsampled low-resolution data, both the pixel-wise error and physical metrics improve, except for the compliance error in seen boundary conditions. Remarkably, for compliance error in seen boundary conditions, the use of upsampled low-resolution data results in predicting almost 200 more feasible designs. These observations suggest that improving the performance of the low-resolution generative model could be a good start for improving the cascading diffusion pipeline.

\subsection{Limitations and Future Work}
While our study demonstrates the effectiveness and generalizability of the cCDM compared to the cGAN model with transfer learning, several challenges remain. One important pitfall of the diffusion model, in contrast to the cGAN model, is its computational inference time. For instance, our low-resolution diffusion model on NVIDIA GeForce RTX 3080 Ti required 4.81 seconds, whereas the high-resolution model took 10.93 seconds to generate a single topology. In comparison, the cGAN model only took 2.09 seconds to produce similar results on the same hardware. Reducing the computational time of diffusion models is actively under investigation~\cite{ma2022} which could help for effectiveness of cCDM.

Another significant limitation of our work is its focus on 2D structural problems. Future research directions could include extending our model to address three-dimensional or more complex problems, thereby broadening its applicability and impact. Additionally, a key challenge in using generative models for topology optimization lies in incorporating objective values and ensuring feasibility without imposing additional computational costs. While some researchers have made progress in this area ~\cite{maze2023diffusion, behzadi2022, giannone2023}, further investigations are needed to streamline this process and minimize associated costs.

Furthermore, future work could explore quantifying the tradeoff between the cost of low-resolution and high-resolution samples, particularly regarding their impact on generative model performance. Understanding this tradeoff could provide information regarding the optimal ratio of sample types, thereby maximizing the effectiveness of generative models for multiresolution tasks.

\section{Conclusions}
We proposed the cCDM to investigate the performance of the diffusion model in the context of the multi-resolution for engineering design problem. We benchmarked our model against cGAN model with transfer learning. Our results demonstrate that the cCDM outperforms the cGAN model in terms of capturing finer details, preserving volume fraction constraints, and minimizing compliance errors in multi-resolution task when there are enough (more than 102 designs) high-resolution training data. Furthermore, we studied the impact of training data size on both models' performance. While both models showed a decrease in performance with reduced training data, the cCDM lost its superiority to cGAN model with transfer learning in limited training data and we found the breakeven point for this change. While we showed that the diffusion model might produce the best pixel-wise performance in low-resolution and high-resolution it does not guarantee the best performance in design metrics. Also, our experiment isolating the effect of the low-resolution generative model on the downstream super resolution tasks showed that meaningful improvements in the low-resolution model can significantly improve the joint accuracy of our cascaded pipeline. In conclusion, our study underscores the potential of cCDM as effective frameworks for multi-resolution inverse design problem tasks offering superior performance and generalizability compared to cGAN model with transfer learning when there exists enough high-resolution training data.

\section*{Acknowledgements}
We acknowledge the support from the National Science Foundation through award \#1943699. We also acknowledge several helpful conversations with Mohammad Mahdi Behzadi that allowed us to accurately replicate the results from their original paper. We also acknowledge  Cashen Diniz for helpful discussions regarding the implementation of diffusion models.  

\bibliographystyle{asmeconf}
\bibliography{references}
\appendix       
\clearpage
\section*{Supplemental Material}
\begin{figure}[hb!]
\centering
\includegraphics[width=\linewidth]{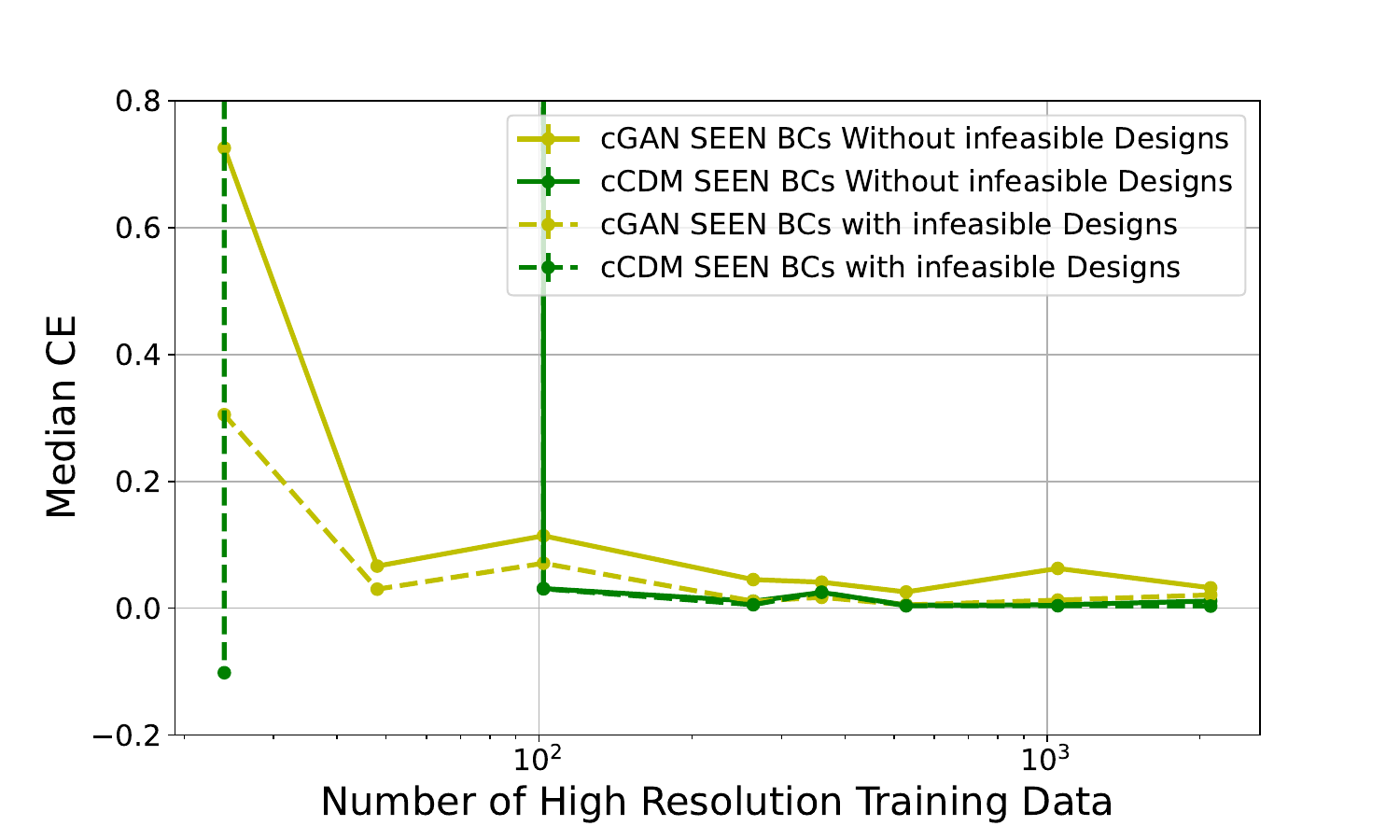}
\caption{Comparison the median of  CE between the cGAN model with transfer learning and the cascaded diffusion models across seen test datasets when the models' prediction includes/excludes infeasible designs.}
\label{CE4.ps}
\end{figure}
\begin{figure}[hb!]
\centering
\includegraphics[width=\linewidth]{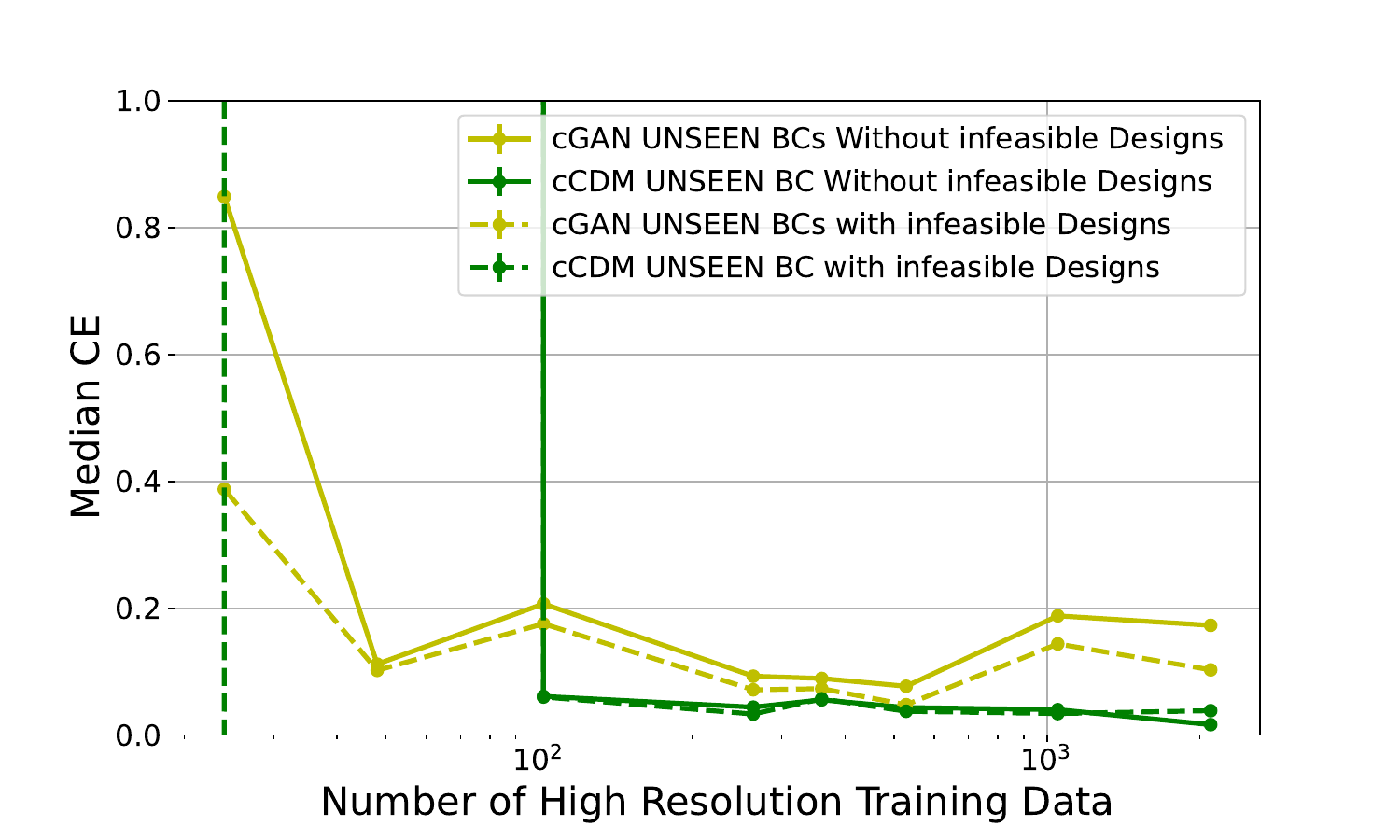}
\caption{Comparison of the median of CE between the cGAN model with transfer learning and the cascaded diffusion models across unseen test datasets when the models' prediction includes/excludes infeasible designs.}
\label{CE3.ps}
\end{figure}

\end{document}